\documentclass[sn-mathphys,Numbered]{sn-jnl}

\usepackage{graphicx}%
\usepackage{multirow}%
\usepackage{amsmath,amssymb,amsfonts}%
\usepackage{amsthm}%
\usepackage{mathrsfs}%
\usepackage[title]{appendix}%
\usepackage{xcolor}%
\usepackage{textcomp}%
\usepackage{manyfoot}%
\usepackage{booktabs}%
\usepackage{algorithm}%
\usepackage{algorithmicx}%
\usepackage{algpseudocode}%
\usepackage{listings}%

\usepackage{microtype}
\usepackage{graphicx}
\usepackage{subfigure}
\usepackage{booktabs}
\usepackage{enumitem}
\usepackage{multirow}
\usepackage{amssymb}
\usepackage{amsmath}
\usepackage{xfrac}
\usepackage{color}
\usepackage{mathtools}
\usepackage{bbm}
\usepackage[center]{caption}
\usepackage{amsthm}

\begin{document}

\title[Better Schedules for Low Precision Training of Deep Neural Networks]{Better Schedules for Low Precision Training of Deep Neural Networks}


\author[1, 2]{\fnm{Cameron R.} \sur{Wolfe}}\email{crw13@rice.edu} 

\author[1]{\fnm{Anastasios} \sur{Kyrillidis}}\email{anastasios@rice.edu}


\affil[1]{\orgdiv{Department of Computer Science}, \orgname{Rice University}, \orgaddress{\street{6100 Main Street}, \city{Houston}, \postcode{77005}, \state{TX}, \country{USA}}}
\affil[2]{Corresponding Author}



\abstract{
Low precision training can significantly reduce the computational overhead of training deep neural networks (DNNs).
Though many such techniques exist, cyclic precision training (CPT), which dynamically adjusts precision throughout training according to a cyclic schedule, achieves particularly impressive improvements in training efficiency, while actually improving DNN performance.
Existing CPT implementations take common learning rate schedules (e.g., cyclical cosine schedules) and use them for low precision training without adequate comparisons to alternative scheduling options.
We define a diverse suite of CPT schedules and analyze their performance across a variety of DNN training regimes, some of which are unexplored in the low precision training literature (e.g., node classification with graph neural networks).
From these experiments, we discover alternative CPT schedules that offer further improvements in training efficiency and model performance, as well as derive a set of best practices for choosing CPT schedules.
Going further, we find that a correlation exists between model performance and training cost, and that changing the underlying CPT schedule can control the tradeoff between these two variables.
To explain the direct correlation between model performance and training cost, we draw a connection between quantized training and critical learning periods, suggesting that aggressive quantization is a form of learning impairment that can permanently damage model performance.
}

\keywords{Efficient training, quantization, hyperparameter schedules
}



\maketitle

\section{Introduction}\label{sec1}

\noindent \textbf{Background.}
DNNs have revolutionized the performance of autonomous systems.
Yet, such gains come at a steep cost---DNN training is computationally burdensome and becoming more so, as the community tends towards larger-scale experiments \citep{brown2020language}.
Though many approaches exist for reducing DNN overhead \citep{li2016pruning, goyal2017accurate}, low precision training has gained recent popularity due to its ability to improve training efficiency with minimal performance impact \citep{fu2020fractrain, fu2021cpt}.

\begin{wrapfigure}{r}{0.6\textwidth} 
  \begin{center} \vspace{-0.4cm}
    \includegraphics[width=0.58\textwidth]{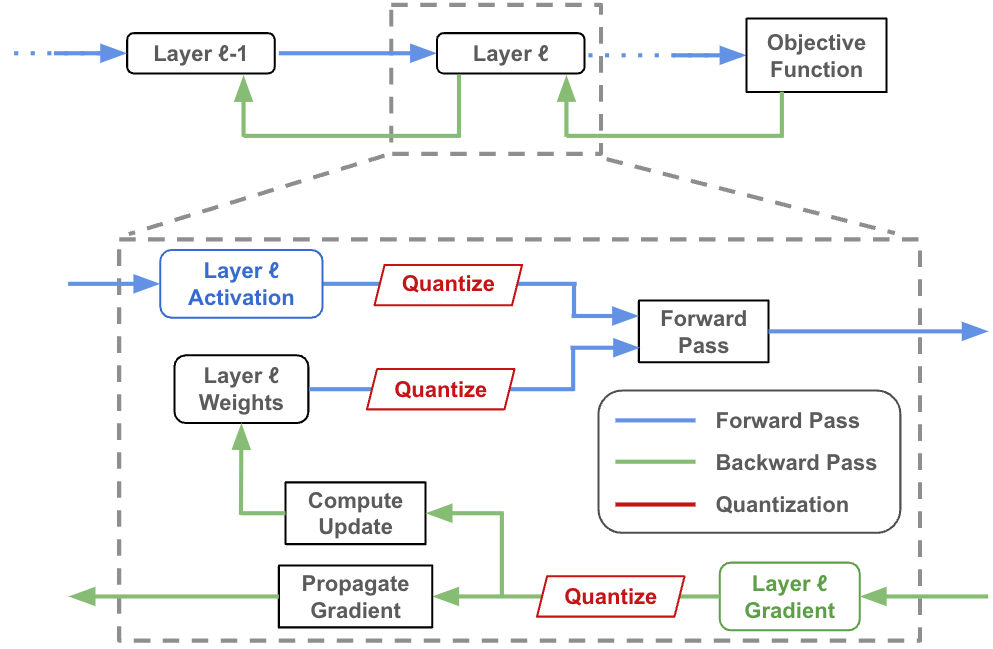}
  \end{center}
  \caption{A depiction of quantized forward/backward pass within a single DNN layer.}
  \label{fig:quant_depict}
  \vspace{-0.6cm}
\end{wrapfigure}

Quantized training approaches use lower precision representations for DNN activations, gradients, and weights during training; see Figure \ref{fig:quant_depict}.
Then, the forward and backward pass can be expedited via (faster) low precision arithmetic, which recent generations of hardware are beginning to support \citep{micikevicius2017mixed}. 
Though implementing quantized training is more nuanced in complex architectures (e.g., batch normalization modules require special treatment \citep{banner2018scalable}), the basic approach remains the same---DNN activations, weights, and gradients are replaced with lower-precision representations during training to reduce the cost of each forward and backward pass.

Early approaches to quantized training adopted a static approach that maintained the same, low level of precision throughout training \citep{banner2018scalable, zhou2016dorefa, yang2020training}.
Although this work was successful in revealing that DNN training is robust to substantial reductions in precision, later work achieved further efficiency gains by dynamically adapting precision along the training trajectory \citep{fu2020fractrain, fu2021cpt}.
Such work $i)$ sets minimum ($q_{\text{min}}$) and maximum ($q_{\text{max}}$) bounds for precision, and $ii)$ varies the precision between these bounds according to some (possibly cyclical) schedule throughout training.
Interestingly, experiments using cyclic precision training (CPT) \citep{fu2021cpt} demonstrated that precision plays a role similar to that of the learning rate in DNN training.

Because state-of-the-art results with DNNs are often achieved using curated hyperparameter schedules \citep{he2016deep, devlin2018bert, smith2018disciplined}, different scheduling options for common hyperparameters (e.g., learning rate and momentum) have been explored extensively \citep{chen2022demon, chen2022rex}.
Despite known similarities between precision and the learning rate, however, no such study has been performed for low precision training---existing approaches to CPT simply adopt common learning rate strategies (i.e., a cyclical cosine schedule \citep{fu2021cpt}) without adequate comparison to alternatives.
As such, \textit{best practices for dynamically varying DNN precision along the training trajectory remain largely unknown}.

\medskip
\noindent \textbf{This work.} 
To close this gap, we provide an extensive empirical study of different CPT variants. 
First, we rigorously define the space of potential CPT schedules by deconstructing such schedules via a three-step process of $i)$ choosing a profile, $ii)$ choosing the number of cycles, and $iii)$ choosing whether cycles are repeated and/or have a ``triangular'' form.
Using this decomposition, we define a suite of ten CPT schedules---including the originally-proposed cyclical cosine schedule \citep{fu2021cpt}---that are analyzed across a variety of domains, including image classification and object detection with convolutional neural networks (CNNs), language modeling with long short-term memory (LSTM) networks \citep{hochreiter1997long}, entailment classification with pre-trained transformers \citep{devlin2018bert}, and graph node classification with graph neural networks (GNNs) \citep{kipf2016semi}.
As a byproduct of this analysis, to the best of the authors' knowledge, we are the first to study quantized training strategies for GNNs.

Going beyond prior work, we discover new CPT variants that further improve DNN training efficiency and performance relative to prior schedules, revealing that exploring a wider variety of schedules for CPT is beneficial. 
More generally, we observe across all experiments that model performance tends to be correlated with the amount of compute used for training, revealing that  manipulating the CPT schedule is a simple tool for controlling the tradeoff between model performance and training efficiency.
Aiming to explain this relationship, we draw a connection between quantized training and critical learning periods, finding that prolonged training at low precisions can impair the training process and cause a permanent performance degradation.
Finally, we leverage these empirical observations to provide best practices for choosing the correct CPT schedule in different practical scenarios.  

\section{Related Work} \label{S:related_work}

\textbf{Neural Network Quantization.} Several works leverage DNN quantization to construct high-performing, efficient DNNs \citep{xu2018dnq, park2020profit, wang2019haq, esser2019learned, bhalgat2020lsq+}.
\cite{jacob2018quantization} studies quantization-aware training strategies to facilitate post-training DNN quantization, \cite{jung2019learning} adopts learnable approaches for DNN quantization, and \cite{wang2019haq} applies adaptive precision to different DNN components (e.g., layers or filters) during inference. 
Binary and ternary DNNs have also been studied \citep{courbariaux2016binarized, li2016ternary}.
For GNNs, a few works have studied post-training quantization \citep{tailor2020degree, feng2020sgquant}, but quantized training of GNNs has not been considered.

\medskip \noindent
\textbf{Quantized Training.} Several works, as in \citep{gupta2015deep, wang2018training, micikevicius2017mixed, banner2018scalable, scao2022bloom}, pioneer low precision DNN training, finding that quantized training at a static, low level yields significant time and energy savings. 
Later work explores adaptive precision throughout training \citep{fu2020fractrain, fu2021cpt}.
Such methods can be applied on top of static quantization schemes and are found to yield gains in efficiency and model performance, by dynamically lowering precision below that of static techniques during training.

\medskip \noindent
\textbf{Critical Learning Periods.} Critical learning periods within deep neural networks describe the early training epochs, during which the network is most sensitive to learning impairments.
Early work on critical learning periods found that neural networks trained over blurry images for a sufficient number of epochs could never recover the accuracy of a network trained normally \citep{achille2018critical}.
Subsequent work studied different forms of training impairments, such as improper regularization and non-i.i.d. data \citep{golatkar2019time, ash2020warm}.
Work on critical learning periods reveals that learning deficits during an initial, sensitive phase of training yield a permanent degradation in performance, no matter the amount of training performed after removing the deficit.

\medskip \noindent \textbf{Hyperparameter Schedules.}
Most state-of-the-art results on deep learning benchmarks use curated hyperparameter schedules \citep{he2016deep, smith2017cyclical}.
Scheduling is commonly used for hyperparameters like the learning rate \citep{loshchilov2016sgdr}, but the general concept is widely-applied within deep learning.
For example, scheduling approaches have been proposed for mini-batch sizes \citep{wu2020multigrid}, image spatial resolution \citep{prog_resize}, the amount of regularization \citep{smith2022general}, and more.
Best practices for hyperparameter scheduling have been established through extensive empirical analysis; e.g., for the learning rate \citep{chen2022rex, smith2017cyclical}, momentum \citep{chen2022demon}, and even batch size and weight decay \citep{smith2018disciplined}.
Our work aims to provide empirical analysis that establishes such best practices for low precision training.

\begin{figure*}
    \centering
    \includegraphics[width=1\linewidth]{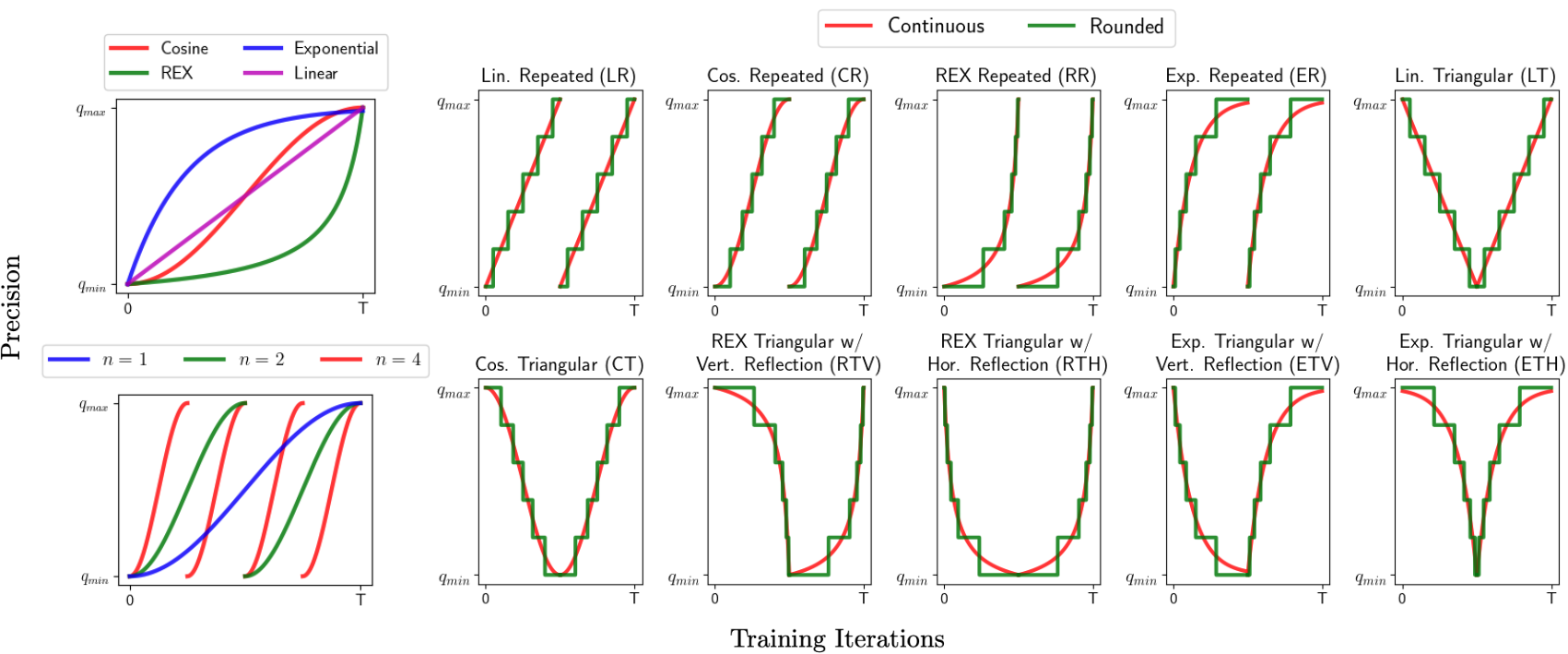} \vspace{-0.4cm}
    \caption{An illustration of profiles and schedules for CPT over $T$ total training iterations. Function profiles are depicted in the upper-left subplot, while the lower-left subplot illustrates the CR schedule with different numbers of cycles $n$. Remaining subplots depict all possible CPT schedules explored in this work---both with and without rounding to the nearest integer---for $n=2$ cycles.} 
    \label{fig:schedules} \vspace{-0.3cm}
\end{figure*}

\section{Precision Schedules for Quantized Training} \label{S:method}
Our goal in exploring different CPT schedules is to $i)$ better understand the impact of such schedules on DNN performance, and $ii)$ study new schedules that offer different levels of reductions in DNN training cost.
We aim to make our analysis comprehensive by considering a wide variety of schedules.
We will now provide a brief overview of CPT and explain how the suite of CPT schedules used in this work is derived.


\subsection{How does CPT work?} \label{S:cpt_explain}
$\texttt{Quantize}$ in Figure \ref{fig:quant_depict} converts data into a lower precision representation, before it is used in the forward or backward pass.
We refer to this lower level of precision as the ``target'' precision.
CPT operates by simply varying this target precision such that each training iteration $t$ has a different target precision $q_t$.
More specifically, CPT varies target precision within the range $[q_{\text{min}}, q_{\text{max}}]$, according to some schedule during training.
We define this schedule as a function $S(\cdot) : \mathbb{N}_{\geq 0} \rightarrow \mathcal{Q}$, where $\mathcal{Q}$ is the set of real-valued numbers in the range $[q_{\text{min}}, q_{\text{max}}]$.
The precision used at iteration $t$ of training, which is always rounded to the nearest integer, is given by $q_t = \texttt{round}(S(t)) \in [q_{\text{min}}, q_{\text{max}}]$.

$q_{\text{max}}$ is selected to match the precision of static, low precision training, ensuring that CPT saves costs by adjusting DNN precision below this static level.
$q_{\text{min}}$ must be discovered via a precision range test (see Section 3.3 in \citep{fu2021cpt}), as DNN training cannot progress when precision is too low.
To stabilize training, cyclic precision is only applied to the forward pass (i.e., activation and weights quantization in Figure \ref{fig:quant_depict}), while the backward pass (i.e., gradient quantization in Figure \ref{fig:quant_depict}) fixes precision at $q_{\text{max}}$ throughout training.

\subsection{Constructing a CPT Schedule} \label{S:sched_construct}

Constructing a low precision training schedule can be decomposed into three steps:\footnote{Related work \citep{chen2022rex} considers a sampling rate for each profile, which controls the frequency of hyperparameter updates. This sampling rate is less pertinent to precision schedules because precision is always rounded to the nearest integer, which makes updates less frequent.}
\begin{itemize}
    \item[1.] Selecting a Profile
    \item[2.] Setting the Number of Cycles
    \item[3.] Selecting Repeated or Triangular Cycles
\end{itemize}
The remainder of this section explains this decomposition and how it is used to derive the suite of CPT schedules explored in this work.


\medskip
\noindent \textbf{Step One: Selecting a Profile.}
Any CPT schedule must have an underlying function (``profile'') according to which precision is varied.
We consider four function profiles---cosine, linear, exponential, and reverse exponential (REX) \citep{chen2022rex}; see top left subplot of Figure \ref{fig:schedules}.
We only consider growth profiles (i.e., functions that increase precision from $q_{\text{min}}$ to $q_{\text{max}}$), because DNN training must end with higher precision to facilitate convergence \citep{fu2021cpt}.
This set of functions characterizes a range of different quantization behaviors that reduce computational cost to varying degrees.

\medskip
\noindent \textbf{Step Two: Setting the Number of Cycles.}
Beyond choosing a profile, each schedule may perform a certain number of cycles, which we denote as $n$, according to this profile during training.
Different choices for $n$ are depicted in the bottom left subplot of Figure \ref{fig:schedules}.
In our experiments, we find that $n=8$ performs consistently well.

\medskip
\noindent \textbf{Step Three: Selecting Repeated or Triangular Cycles.}
The proposed profiles can be used to produce many different schedules.
Each cycle may repeatedly increase precision from $q_{\text{min}}$ to $q_{\text{max}}$; see ``repeated'' subplots in Figure \ref{fig:schedules}.
In contrast, a schedule can be made ``triangular'' by reflecting the profile within every other cycle \citep{fu2021cpt, smith2018disciplined, smith2017cyclical}; see top right and bottom repeated subplots in Figure \ref{fig:schedules}.
Assuming $n$ is even and that all schedules end with a precision of $q_{\text{max}}$ to facilitate model convergence, we can construct triangular schedules by reflecting all odd-numbered cycles, leading adjacent cycles to vary precision in opposite directions; see LT and CT subplots of Figure \ref{fig:schedules}.
REX and Exponential function profiles can be reflected either vertically or horizontally\footnote{Cosine and linear function profiles are symmetric, which causes horizontal and vertical reflections to be identical.}, producing two different triangular schedule variants; see the RTV, RTH, ETV and ETH subplots within Figure \ref{fig:schedules}.

\medskip
\noindent \textbf{Suite of CPT Schedules.}
The full set of CPT schedules we consider is composed of all repeated and triangular variants of the linear, cosine, REX, and exponential function profiles. 
Notably, this set includes the original cyclical cosine schedule proposed for CPT (CR in Figure \ref{fig:schedules}) \citep{fu2021cpt}.
Both repeated and triangular schedules can be repeated for any (even) number of cycles, though we set $n=8$ in most experiments.
The training efficiency of each schedule, relative to the others, does not change (assuming all schedules use the same setting of $q_{\text{min}}$ and $q_{\text{max}}$ for comparable experiments).
With this in mind, we organize our suite of CPT schedules into three groups, which we refer to as Small, Medium, and Large based upon their provided reductions in training cost.
\begin{itemize}[leftmargin=*]
    \item Group I (Large): RR, RTH
    \item Group II (Medium): LR, LT, CR, CT, RTV, ETV 
    \item Group III (Small): ER, ETH
\end{itemize}
To ease readability, we will use these groups to refer to different CPT schedules throughout the remainder of the text.

\section{Experiments} \label{S:experiments}
We perform experiments across a variety of domains, including image recognition (i.e., image classification and object detection), node classification, and language understanding (i.e., language modeling and entailment classification).
Results are evaluated based upon model performance and the computational cost of training.
For each experimental domain, we first provide details about the setup, then present and analyze results.
The section concludes with an empirically-derived set of practical suggestions for selecting a CPT schedule.

\subsection{Preliminaries}
We set $q_{\text{max}} \in \{6, 8 \}$ and $n=8$.
$q_{\text{min}}$ is derived for each model-dataset pair using a precision range test.
Our baseline, inspired by SBM \citep{banner2018scalable}\footnote{Prior work \citep{fu2021cpt} shows that SBM outperforms other static techniques for quantized training.},  maintains a static precision of $q_{\text{max}}$ throughout training.
To quantify the computational cost of training, we report the effective number of bit operations \citep{zhou2016dorefa}, which is proportional to the total number of bit operations during training.
This quantity can be computed as:
\begin{align*}
     \texttt{BitOps} = \texttt{FLOP}_{a \times b} \cdot \left (\sfrac{\texttt{Bit}_a}{32} \right) \cdot \left ( \sfrac{\texttt{Bit}_b}{32}  \right)
\end{align*}
for a dot product between $a$ and $b$ with precisions $\texttt{Bit}_a$ and $\texttt{Bit}_b$, respectively, and total number of floating point operations $\texttt{FLOP}_{a\times b}$.

\medskip \noindent \textbf{Implementation.}
All image recognition experiments are implemented using PyTorch and Torchvision \citep{paszke2017automatic}.
We implement GNN training using the Deep Graph Library \citep{wang2019deep}, which provides a message passing framework for the communication of data within a graph, and PyTorch.
For language modeling experiments, we base our implementation upon publicly available repositories for both language modeling and entailment classification \citep{zaremba2014recurrent, wolf2020transformers}.
Experiments are run using a single Nvidia 3090 GPU.
Because current generations of GPUs do not support arbitrary precision levels \citep{micikevicius2017mixed}, we adopt the approach of prior work \citep{fu2020fractrain, fu2021cpt} and simulate low precision training with different bit widths by clipping activation and gradient information beyond the current precision level $q_t$ at every iteration of training.

\begin{figure}[!t]
    \centering
    \includegraphics[width=.9\linewidth]{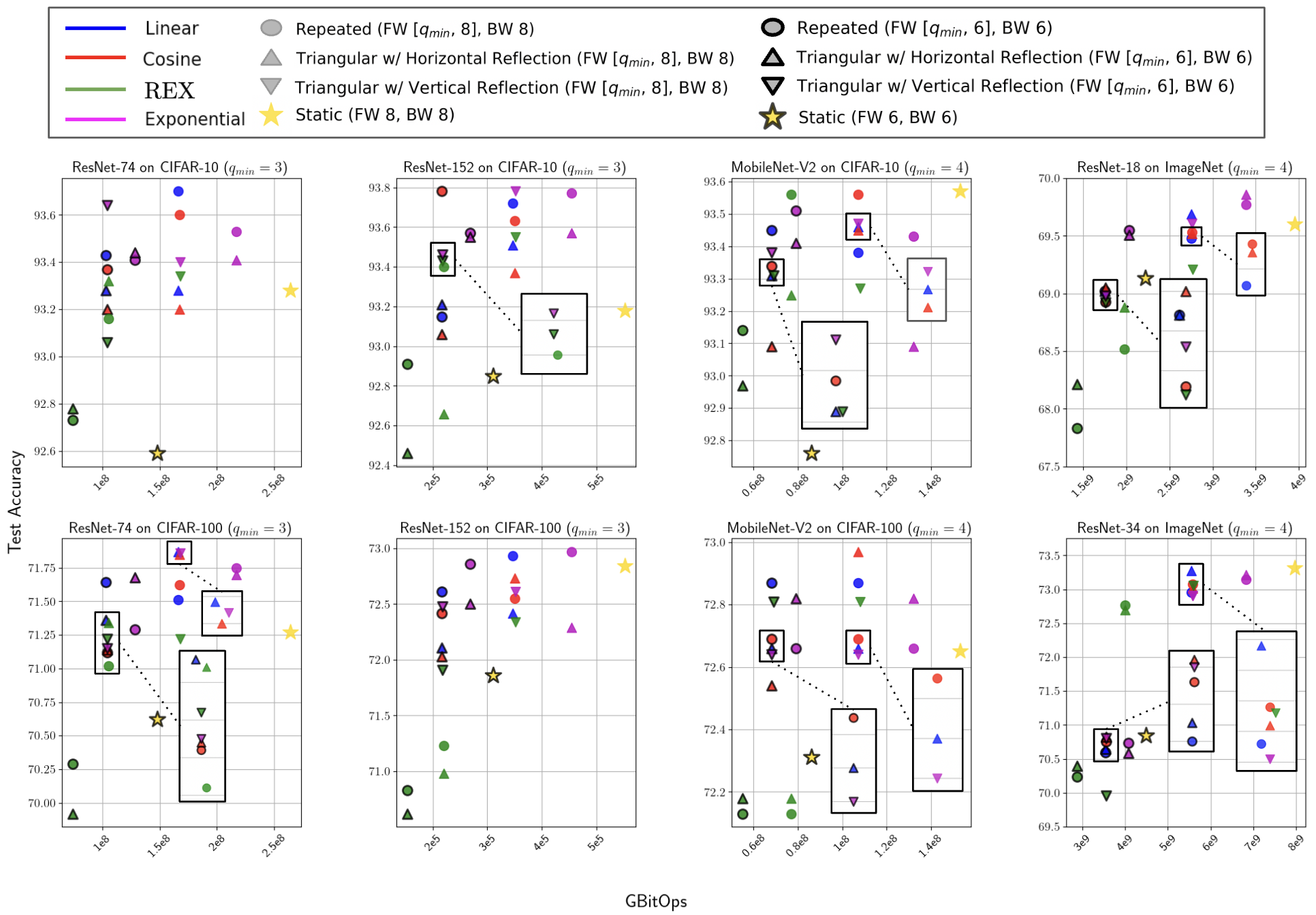}
    \vspace{-0.1cm}
    \caption{Results of CPT experiments on CIFAR-10/100 and ImageNet. Colors represent profiles, while shapes distinguish repeated or triangular schedules. Experiments are run with $q_{\text{max}} \in \{6, 8\}$, distinguished by a dark outline around a shape. Future figures adopt the same scheme of colors and shapes.}
    \label{fig:img_classif_res}
    \vspace{-0.5cm}
\end{figure}

\subsection{Image Recognition} \label{S:img_classif}
We evaluate the proposed CPT schedules on both image classification and object detection tasks. 
We train ResNet-74, ResNet-152, and MobileNet-V2 architectures on the CIFAR-10/100 datasets \citep{he2016deep, sandler2018mobilenetv2}, following the settings of \citep{wang2018skipnet}.
Additionally, we adopt the settings of \citep{he2016deep} and perform experiments with ResNet-18 and ResNet-34 architectures on ImageNet.
Object detection experiments are performed on the PascalVOC dataset \citep{pascal-voc-2012} and adopt a RetinaNet architecture \citep{lin2017focal} with an ImageNet pre-trained ResNet-18 backbone \citep{he2016deep}.
Performance is reported in terms of test accuracy and mean average precision (mAP) for image classification and object detection experiments, respectively. 
For all experiments, we report performance as an average across three separate trials and all hyperparameters are tuned using a hold-out validation set.

\medskip
\noindent \textbf{CIFAR-10/100.}
We adopt the settings of \citep{wang2018skipnet} for training.
All models are trained for a total of $64K$ iterations with a batch size of 128 using standard crop and flip data augmentations.
We use a SGDM optimizer with momentum of 0.9.
Training begins with a learning rate of 0.1, and the learning rate is decayed by $10\times$ after 50\% and 75\% of training iterations. 
Weight decay is set to $1 \times 10^{-4}$, and we use $q_{\text{min}} = 3$ (i.e., discovered using a precision range test) and $q_{\text{max}} \in \{6, 8\}$.
All CPT variants are tested with $n=8$.

Results of experiments with CPT on CIFAR-10 and CIFAR-100 are shown in Figure \ref{fig:img_classif_res}.
Despite using less training compute, CPT variants tend to outperform static baselines in nearly all cases.
This finding aligns with prior work \citep{fu2021cpt} but goes further by demonstrating that \textit{the benefit of CPT is not specific to any particular precision schedule.}
Compared to CPT with the originally-proposed CR schedule, our large schedules achieve further reductions in training compute, but slightly degrade accuracy.
Small and medium schedules tend to both reduce training cost and improve accuracy.

Across all experiments, we see that a correlation exists between model performance and training compute, irrespective of model depth or architecture.
This interesting finding reveals that \textit{the choice of CPT schedule is a mechanism that controls the trade off between model performance and training compute}.
Compared to most hyperparameter settings (e.g., the learning rate) that only impact model performance, the choice of CPT schedule is quite nuanced due to its joint impact on training efficiency and performance.

\medskip
\noindent \textbf{ImageNet.} We consider the ILSVRC2012 version of ImageNet with 1K total classes. 
The settings of \citep{he2016deep} are used: 
Models are trained for a total of 90 epochs with a batch size of 256 using standard crop and flip data augmentations. 
We use a SGDM optimizer with momentum of 0.9.
An initial learning rate of 0.1 is adopted, and this learning rate is decayed by a factor of $10\times$  after 50\% and 75\% of total epochs.
Weight decay is set to $1 \times 10^{-5}$, and we use $q_{\text{min}} = 4$ and $q_{\text{max}} \in \{6, 8\}$.
All CPT variants are tested with $n=8$.

As shown in Figure \ref{fig:img_classif_res}, CPT schedules that significantly reduce training compute cause a noticeable performance deterioration relative to static baselines on ImageNet.
For example, large schedules cause a $0.5-1.5\%$ drop in accuracy relative to static baselines across different architectures and settings of $q_{\text{max}}$.
With the larger ResNet-34 architecture, we see that more aggressive quantization is especially detrimental to performance; e.g., setting $q_{\text{max}}=6$ significantly deteriorates performance in both CPT and baseline experiments.

\begin{wrapfigure}{r}{0.5\textwidth} 
  \begin{center} \vspace{-0.8cm}
    \includegraphics[width=0.48\textwidth]{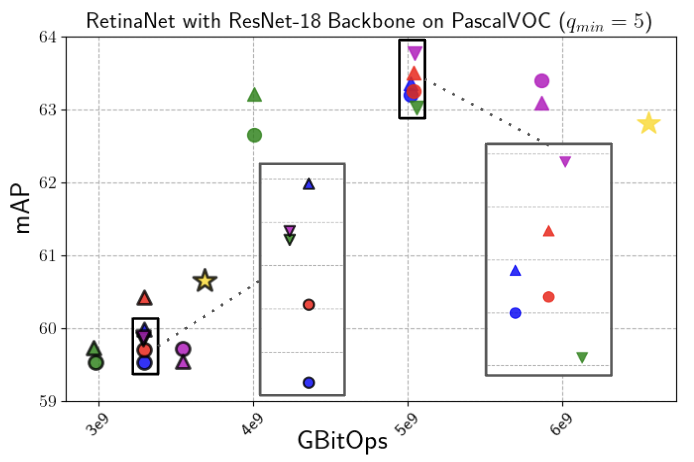}
  \end{center}
  \caption{Results of CPT experiments on PascaVOC. The same coloring scheme is adopted from Figure \ref{fig:img_classif_res}.}
  \label{fig:obj_det}
  \vspace{-0.8cm}
\end{wrapfigure}

Though we still observe a correlation between performance and training compute, larger-scale image classification experiments seem to be more sensitive to the level of training quantization---aggressive quantization noticeably deteriorates model performance.
By adopting our small schedules that quantize more conservatively, however, we exceed baseline performance at a reduced training cost.
In comparison, medium schedules---including the originally-proposed CR schedule---perform similarly to or worse than baseline models.

\medskip
\noindent \textbf{Pascal VOC.} 
We consider the Pascal VOC 2012 dataset for both training and testing \citep{pascal-voc-2012}.
Prior to training on Pascal VOC, the ResNet backbone of each RetinaNet model is pre-trained on the ILSVRC2012 dataset. 
Models are trained for 120 total epochs with a batch size of 4 on Pascal VOC.  
We do not perform any data augmentation, though images are normalized and resized before being passed as input.
We adopt an Adam optimizer \citep{kingma2014adam} for training with a fixed learning rate of $1 \times 10^{-5}$ throughout training.
The model is trained using a focal loss that matches the settings of \citep{lin2017focal}.
We use $q_{\text{min}} = 5$, $q_{\text{max}} \in \{6, 8\}$, and $n=8$.

The results of CPT experiments on PascalVOC are shown in Figure \ref{fig:obj_det}.
Here, we see that setting $q_{\text{max}} = 6$ yields a clear performance deterioration in both baseline and CPT experiments, indicating that training on PascalVOC is sensitive to quantization.
When $q_{\text{max}} = 8$, however, \textit{all CPT variants either match or exceed the performance of static baselinesm while reducing the cost of training}.
For example, the best medium schedule (ETV) yeilds an absolute improvement of 1.02 mAP with a 25\% reduction in training cost.
The performance of all CPT schedules is roughly comparable when $q_{\text{max}} = 8$.

\subsection{Node Classification} \label{S:node_classif}
Graph node classification experiments are conducted with CPT on OGBN-Ariv and OGBN-Products datasets \citep{hu2020open}.
On OGBN-Arxiv, we use a regular GNN architecture \citep{kipf2016semi} and consider the full graph during each training iteration.
Experiments on OGBN-Products use a GraphSAGE architecture \citep{hamilton2017inductive} with random neighbor sampling.
Given that we are the first to explore low precision training within this domain, we first define low precision training with respect to the GNN architecture and identify the unique aspects of training GNNs with CPT.

\medskip \noindent \textbf{Low Precision GNN Training.}
Consider a graph $\mathcal{G}$ comprised of $e$ edges and $n$ nodes with $d$-dimensional features $\mathbf{X} \in \mathbb{R}^{n \times d}$.
The output $\mathbf{Y} \in \mathbb{R}^{n \times d_{L}}$ of a GNN is given by $\mathbf{Y} = \Psi_{\mathcal{G}}(\mathbf{X}; \boldsymbol{\Theta})$, where $\Psi_{\mathcal{G}}$ is a GNN architecture with $L$ layers and (trainable) parameters $\boldsymbol{\Theta}$.  
Given $\mathbf{H}_0 = \mathbf{X}$, we have $\mathbf{Y} = \Psi_{\mathcal{G}}(\mathbf{X}; \boldsymbol{\Theta}) = \mathbf{H}_L$, with $\ell$-th layer GNN representations
\begin{align}\label{gcn_forward}
    \mathbf{H}_{\ell} = \sigma(\bar{\mathbf{A}} \, \mathbf{H}_{\ell-1} \, \boldsymbol{\Theta}_{\ell - 1}).
\end{align}
where $\sigma(\cdot)$ is an elementwise activation function (e.g., ReLU), $\bar{\mathbf{A}}$ is the degree-normalized adjacency matrix of $\mathcal{G}$ with added self-loops, and the trainable parameters $\boldsymbol{\Theta} = \{\boldsymbol{\Theta}_\ell\}_{\ell=0}^{L-1}$ have dimensions $\boldsymbol{\Theta}_\ell \in \mathbb{R}^{d_{\ell} \times d_{\ell+1}}$ with $d_0  = d$ and $d_L = d'$.
The GNN architecture is similar to a feed-forward neural network with an added node feature \textbf{aggregation step}, characterized by the left-multiplication of node features by $\bar{\mathbf{A}}$ in \eqref{gcn_forward}, at each layer.

To apply quantized training GNNs, we must determine whether this aggregation step can be performed in low precision.
We compare two strategies---$\texttt{FP-Agg}$ and $\texttt{Q-Agg}$. 
$\texttt{Q-Agg}$ indicates that the aggregation process is quantized---either according to a CPT schedule or a fixed precision level---within the GNN's forward pass.
In contrast, $\texttt{FP-Agg}$ always performs aggregation in full precision, no matter the precision level used in the rest of the network.

\medskip
\noindent \textbf{Node classification details.}
Before we present our findings, let us first provide the details of our experiments.
For the OGBN-Arxiv dataset, we adopt a normal GNN model \citep{kipf2016semi} with three layers and a hidden dimension of $512$.
Training on OGBN-Arxiv proceeds for 1000 epochs using an Adam optimizer \citep{kingma2014adam}.
The learning rate begins at an initial value of $10^{-3}$ and decays by a factor of $10\times$ over the course of training using cosine annealing.
All CPT scheduling variants described in Section \ref{S:sched_construct} are tested using $n=8$.
We set $q_{\text{min}} = 3$ (i.e., discovered using a precision range test \citep{fu2021cpt}) and consider $q_{\text{max}} \in \{6, 8\}$.

For the OGBN-Products dataset, we use a two-layer GraphSAGE \citep{hamilton2017inductive} model with a hidden dimension of $512$.
To make training computationally tractable over this large graph, we adopt random neighbor sampling with a neighborhood size of $32$.
We find that validation accuracy plateaus beyond a neighborhood of this size. 
Training proceeds for 80 epochs using an Adam optimizer \citep{kingma2014adam}.
The learning rate begins at an initial value of $3 \times 10^{-4}$ and decays by a factor of $10\times$ over the course of training using cosine annealing.
Again, we test all CPT scheduling variants with $n=8$, $q_{\text{min}} = 3$ (i.e., discovered using a precision range test), and $q_{\text{max}} \in \{6, 8\}$.

\begin{figure}
    \centering
    \includegraphics[width=0.8\linewidth]{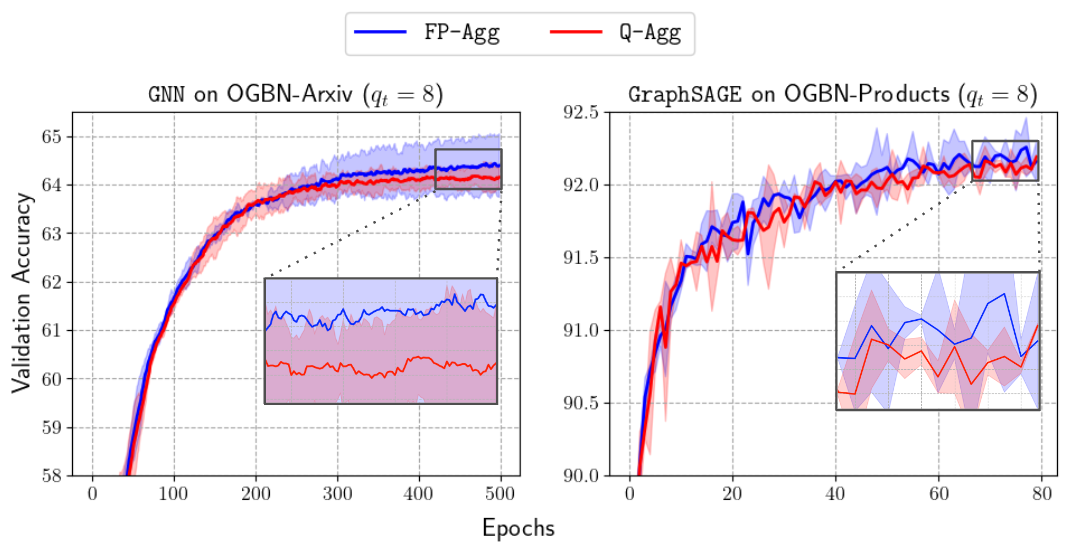}
    \caption{Validation accuracy of GNN and GraphSAGE models trained on OGBN-Arxiv and OGBN-Products using $\texttt{Q-Agg}$ or $\texttt{FP-Agg}$ and $q_{\text{max}} = q_t = 8$.}
    \label{fig:agg_quant}
\end{figure}

\medskip \noindent \textbf{Is aggregation robust to low precision?}
We perform experiments on OGBN-Arxiv and OGBN-Products to compare $\texttt{FP-Agg}$ and $\texttt{Q-Agg}$ strategies; see Figure \ref{fig:agg_quant}.
For these experiments, both strategies adopt a precision level of $q_t = q_{\text{max}} = 8$ throughout the training process.

On OGBN-Arxiv, utilizing the $\texttt{Q-Agg}$ strategy yields a slight, but consistent, degradation in performance.
The GNN trained using $\texttt{FP-Agg}$ achieves a 0.5\% absolute improvement in accuracy compared to a GNN trained with $\texttt{Q-Agg}$.
This difference in performance is less pronounced on OGBN-products---$\texttt{FP-Agg}$ and $\texttt{Q-Agg}$ strategies train GraphSAGE models to similar accuracy.\footnote{The lesser impact of $\texttt{Q-Agg}$ on OGBN-Products is due to the use of neighborhood sampling. The aggregation process computes a sum over all neighboring features, which can generate large, numerically unstable components unless the sum is truncated to a smaller, fixed number of neighboring features.}

\begin{figure}
    \centering
    \includegraphics[width=0.85\linewidth]{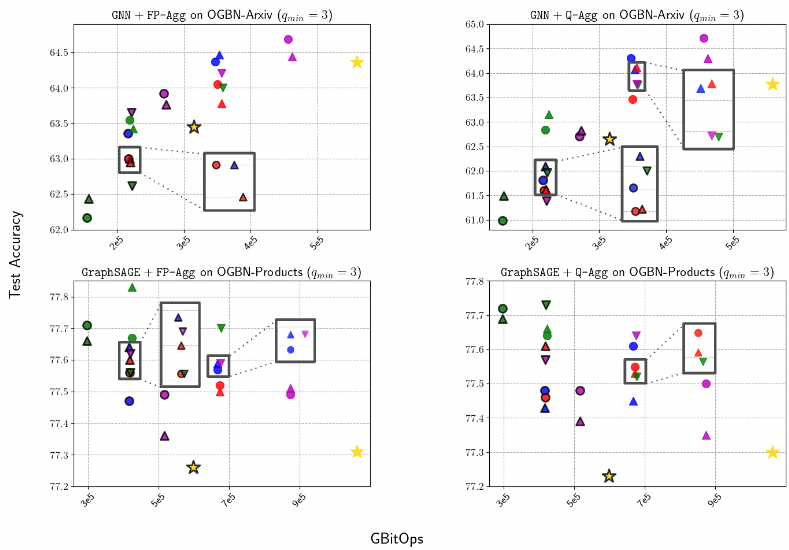}
    \caption{GNN and GraphSAGE test accuracy on OGBN-Arxiv and OGBN-Products. The coloring scheme is adopted from Figure \ref{fig:img_classif_res}.}
    \label{fig:gnn_quant_accs}
\end{figure}

The aggregation process is a negligible portion of the GNN's forward pass.
However, performing aggregation in low precision could greatly benefit communication efficiency in model-parallel training scenarios that pipeline the GNN's forward pass across multiple compute sites \citep{wan2022pipegcn}.
Though we do not consider this case in our work, we analyze both $\texttt{FP-Agg}$ and $\texttt{Q-Agg}$ strategies within the remainder of experiments due to this potential benefit of $\texttt{Q-Agg}$ and its similar level of performance relative to $\texttt{FP-Agg}$.

\medskip \noindent \textbf{OGBN-Arxiv.} 
Results for the GNN model trained with different CPT schedules on OGBN-Arxiv are depicted in Figure \ref{fig:gnn_quant_accs}.
Similarly to experiments on image recognition, we observe a clear relationship between GNN test accuracy and the amount of compute used during training---schedules that perform aggressive quantization tend to be slightly outperformed by those quantizing more modestly.
For example, for both $\texttt{FP-Agg}$ and $\texttt{Q-Agg}$, models trained with CPT using large schedules reach an accuracy 1.0-1.5\% below that of baseline models.

On the other hand, GNN's trained with medium schedules tend to match baseline performance in most cases, while models trained with small schedules outperform baselines in all cases.
Node classification seems to be a complex task that is potentially sensitive to quantized training.
However, by using alternative CPT schedules (i.e., medium or small schedules), we can achieve significant reductions in training compute, while actually improving the performance of the underlying model for both $\texttt{FP-Agg}$ and $\texttt{Q-Agg}$ CPT variants.

\medskip \noindent \textbf{OGBN-Products} 
Compared to experiments on OGBN-Arxiv, GraphSAGE training on OGBN-Products seems to be less sensitive to quantization; see Figure \ref{fig:gnn_quant_accs}.
Nearly all CPT schedules considered yield models that achieve better performance than the baseline models.
For example, using both $\texttt{FP-Agg}$ and $\texttt{Q-Agg}$, large schedules reduce the amount of training compute by $>2\times$, while achieving a 0.3\% to 0.5\% absolute improvement in test accuracy relative to baselines. 
This improvement in test accuracy is standard across nearly all CPT schedules tested on OGBN-Products, indicating that this setting is relatively robust to low precision training.
Significant reductions in training cost can be achieved---without damaging model performance---by using aggressive (large) quantization schedules with CPT.

\subsection{Language Understanding} \label{S:lang_mod}
\begin{wrapfigure}{r}{0.6\textwidth} 
  \begin{center} \vspace{-1cm}
    \includegraphics[width=0.58\textwidth]{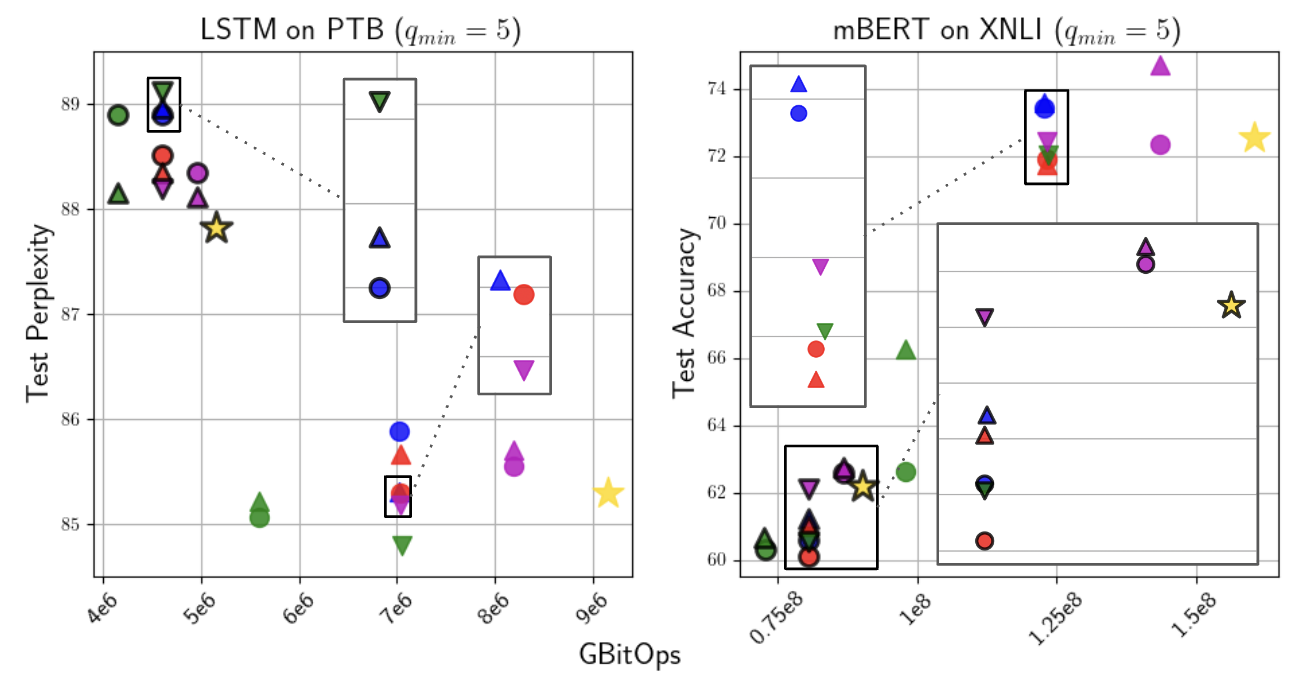}
  \end{center}
  \caption{LSTM and mBERT results on Penn Treebank and the XNLI corpus, respectively. The coloring scheme is adopted from Figure \ref{fig:img_classif_res}.}
  \label{fig:lang_quant}
  \vspace{-0.6cm}
\end{wrapfigure}

We perform language modeling experiments with CPT using LSTM networks on the Penn Treebank dataset.
In these experiments, a one-layer LSTM model \citep{hochreiter1997long} is used to perform word-level language modeling following the settings of \citep{zaremba2014recurrent}, and we evaluate performance in terms of perplexity.
Additionally, we fine-tune a pre-trained, multilingual BERT (mBERT) \citep{devlin2018bert} model (i.e., based on BERT-base-cased) with CPT on the Cross-lingual NLI (XNLI) corpus \citep{conneau2018xnli}, where performance is measured in terms of test accuracy.
In both settings, we report performance as an average across three trials and tune hyperparameters using grid search over a hold-out validation set. 
All results are illustrated within Figure \ref{fig:lang_quant}.

\medskip
\noindent \textbf{Penn Treebank.} 
Experiments follow the setup of \citep{zaremba2014recurrent}.
We use a one-layer LSTM \citep{hochreiter1997long} model with a hidden dimension of 800.
Dropout regularization with $p=0.5$ is applied to the final output layer of the LSTM. 
Models are trained for 40 total epochs using a batch size of 20. 
All training occurs over sequences of length 35 that have been sampled from the Penn Treebank dataset.
Training begins with a learning rate of 20, and the learning rate is divided by five each time the validation accuracy does not improve between epochs.
Throughout training we clip gradients with a maximum norm of 0.25.
For CPT, we adopt $q_{\text{min}} = 5$, $q_{\text{max}} \in \{6, 8\}$, and $n=2$.

As shown in Figure \ref{fig:lang_quant}, the language modeling setting is sensitive to quantization, as revealed by the clear degradation in model performance when $q_{\text{max}}=6$.
Similarly to experiments with PascalVOC in Section \ref{S:img_classif}, however, all CPT variants perform well when $q_{\text{max}}=8$. 
More specifically, we can reduce training cost from $>9$ GBitOps to $5.5$ GBitOps, while improving model perplexity by leveraging large CPT schedules.
In comparison, the original CR schedule matches baseline perplexity at a cost of $7$ GBitOps, again showing that \textit{CPT performance can be improved by exploring alternative schedules within our proposed suite.}

\medskip
\noindent \textbf{XNLI.} 
Experiments follow the settings of a fine-tuning example scripts within the HuggingFace transformers code repository \citep{wolf2020transformers, xnli_script}.
We use the BERT-base-cased-multilingual pre-trained model.
Fine-tuning progresses for 2 total epochs with a batch size of 64 and a sequence length of 128. 
The initial learning rate is chosen from the set $\{5\times10^{-6}, 1\times10^{-5}, 5\times10^{-5}\}$ and is derived separately for each baseline model and CPT scheduling variant using a hold-out validation set.
The learning rate is decayed linearly by $10\times$ throughout fine-tuning.
For CPT, we use $q_{\text{min}}=5$, $q_{\text{max}} \in \{6, 8\}$, and $n=2$.
Notable, we adopt a smaller value of $n$ here because training on proceeds for 2 epochs, and we found that $n \in \{1, 2\}$ perform similarly. 

In experiments with mBERT in Figure \ref{fig:lang_quant}, we again see deteriorated performance when $q_{\text{max}}=6$.
When $q_{\text{max}}=8$, we see a clear correlation between training compute and test accuracy.
For example, large schedules, which require the fewest number of effective bit operations, perform significantly worse than other scheduling variants, while medium schedules tend to match baseline performance at a 25\% reduction in compute costs.
Most interestingly, large schedules (e.g., ETH) improve upon baseline performance by as much as 2\% in absolute test accuracy and still reduce the total cost of training.

\subsection{Best Practices for CPT}
Practitioners will often be faced with choosing a single CPT schedule for training their model.
There is no one CPT schedule that is ``best'' for every domain.
Though the choice of CPT schedule is dependent upon several factors, we can provide the following insight for choosing the most appropriate schedule for a given application.
\begin{itemize}[leftmargin=*]
    \item \textbf{Minimizing training cost:} small schedules yield the largest efficiency gains (but may degrade model performance).
    \item \textbf{Maximizing model performance:} large schedules consistently match or exceed baseline accuracy, even in large-scale experiments.
    \item \textbf{Finding a balance:} medium schedules consistently reduce training cost while maintaining reasonable performance.
\end{itemize}
Such recommendations are reflective of the empirical correlation we observe between model performance and training cost in the majority of domains.
In most cases, as expected, more aggressive quantization during training will come at the cost of slightly reduced model performance.

\section{Connection to Critical Learning Periods} \label{S:crit_learn}
To better understand why we observe a direct relationship between model performance and total cost of training, we draw a connection between low precision training and critical learning periods in deep networks \citep{achille2018critical}.
In particular, we show via experiments across multiple domains that low precision training is a form of training impairment that can cause permanent damage to network performance if applied too aggressively during the early, critical phase of learning. 

\subsection{Is low precision training a learning impairment?}
We consider three settings---image classification with ResNet-74 on CIFAR-10, image classification with ResNet-18 on ImageNet, and node classification with GNNs on OGBN-Arxiv.
All hyperparameter settings match those outlined in Sections \ref{S:img_classif} and \ref{S:node_classif}.\footnote{The manner in which learning rate decay is performed could impact the final result of critical learning period experiments \citep{achille2018critical}.
We test multiple learning rate decay strategies and find that they perform similarly. As such, we adopt a simple schedule that decays the learning rate normally throughout training.}
Two types of experiments are performed:
\begin{itemize}[leftmargin=*]
    \item Train with low precision for $R$ epochs or iterations at the beginning of training ($q_t = q_{\text{min}})$ for $t < R$), then use high precision for the remainder of training ($q_t = q_{\text{max}}$ for $t > R$). 
    \item Train with low precision ($q_t = q_{\text{min}}$) during a span of training iterations and adopt a high precision level ($q_t = q_{\text{max}}$) outside of this span (i.e., ``probing'').
\end{itemize}
The goal of these experiments is to determine $i)$ if low precision training impairs a neural network's learning process and $ii)$ whether this deficit or impairment is specific to the early, critical phase of learning.

\medskip \noindent \textbf{Node Classification.}
Critical learning experiments with node classification on OGBN-Arxiv match the hyperparameter and architecture settings described above.
In the first group of experiments, we begin by training GNNs with low precision for $R$ epochs ($q_t = q_{\text{min}} = 3$ for $t < R$), where several different values of $R \in \{0, 100, 200, \dots, 1000\}$ are tested.
After this initial period of low precision training, the GNN is further trained using higher precision ($q_t = q_{\text{max}} = 8$) for 1000 epochs (i.e., the normal training duration). 

In the second group of experiments, we train the GNN for 2000 total epochs.
During training, we perform low precision training ($q_t = q_{\text{min}} = 3$) for a total of 500 epochs.
These 500 epochs of low precision training are placed at different points in the training process.
In particular, we consider the following low precision training windows: $[0, 500]$, $[100, 600]$, $[200, 700]$, $[300, 800]$, $[400, 900]$.\footnote{Here, each number represents an epoch. The window $[100, 600]$ means that low precision training was performed between epochs 100 and 600.}
Low precision training is performed in these windows, and normal precision ($q_t = q_{\text{max}} = 8$) is adopted outside of the windows. 
We test each window with a separate experiment. 

\begin{figure}[!t]
    \centering
    \includegraphics[width=0.8\linewidth]{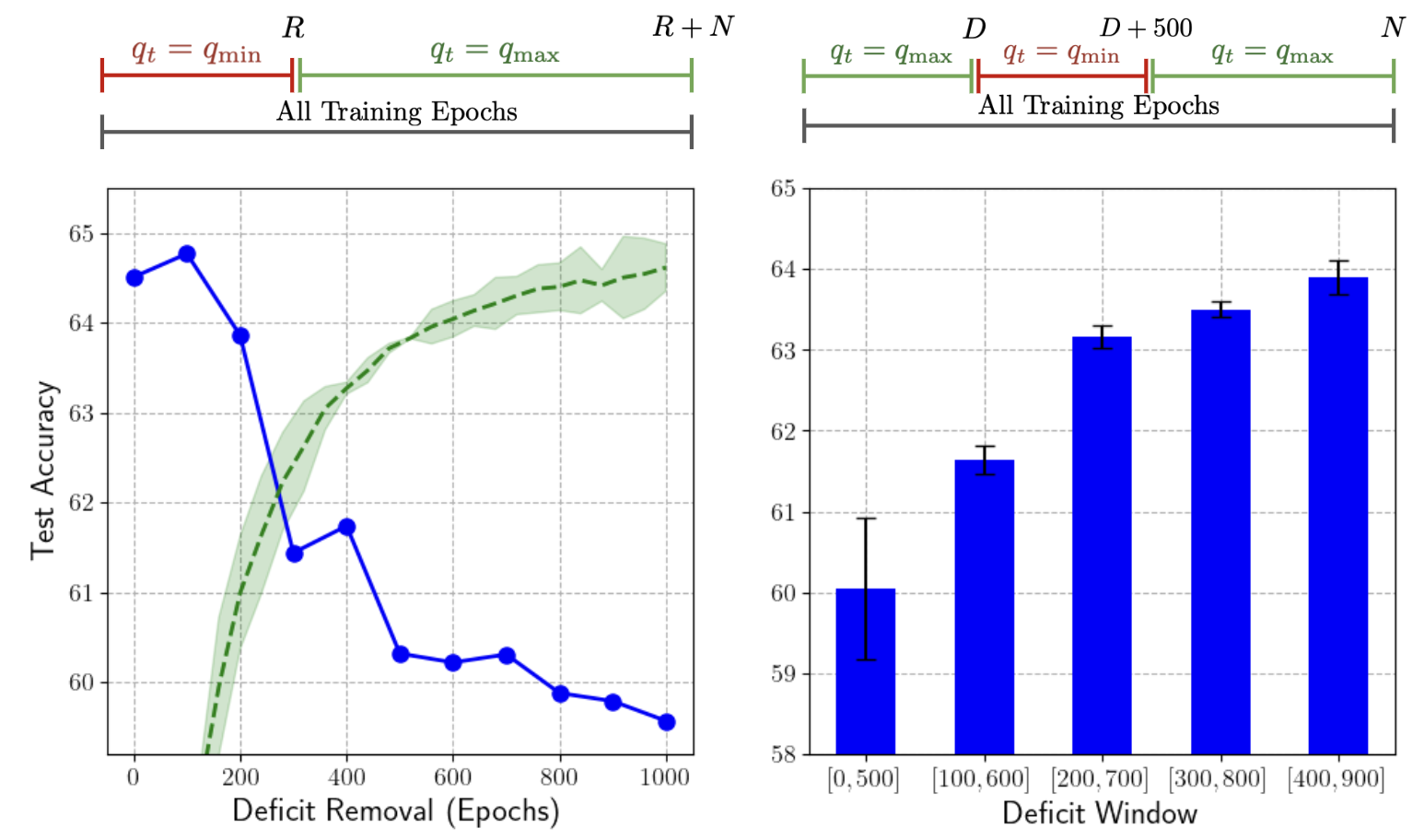}
    \caption{Test accuracy of GNNs trained on OGBN-Arxiv with different forms of learning impairments. (left-blue) Final accuracy of GNNs that are trained using $q_t = 3$ for $R$ epochs, then trained normally with $q_t = 8$ for 1000 epochs. (left-green) Per-epoch test accuracy of a GNN trained normally with $q_t = 8$. (right) Final test accuracy of GNNs trained for a total 1000 epochs using $q_t = 8$, where a 500 epoch window using a lower precision of $q_t = 3$ is placed at different points within the training process.}
    \label{fig:gnn_clp}
    \vspace{-0.2cm}
\end{figure}

The results of critical learning period experiments with OGBN-Arxiv are displayed in Figure \ref{fig:gnn_clp}. 
As shown in the leftmost subplot, the model's test accuracy deteriorates smoothly as the value of $R$ increases, indicating that adopting low precision for a sufficient duration of time at the beginning of training can permanently damage model performance.
The most significant deterioration in GNN test accuracy occurs when lower precision is maintained throughout the early training epochs, during which model accuracy improves the most (i.e., see the green curve in the left subplot of Figure \ref{fig:gnn_clp}). 
If this deficit is removed quickly, model performance does not deteriorate as drastically.

In probing experiments, shown in the right subplot of Figure \ref{fig:gnn_clp}, we see that applying window of low precision training yields the most noticeable accuracy deterioration at the beginning of the training process.
Such a finding reveals that the performance deterioration associated with sufficiently-long periods of low precision training seems to be specific to the early, critical period of training.
\textit{Such a finding provides insight as to why CPT deteriorates model performance when aggressive quantization schedules are adopted}.

\begin{table}[!t]
\centering
\begin{small}
\setlength{\tabcolsep}{5pt}
\begin{tabular}{cll}
\toprule
Setting & Deficit Window & Test Accuracy \\ \midrule
\multirow{9}{*}{ResNet-74 on CIFAR-10} & None & 92.23 $\pm$ 0.09 \\
& [0, 16K] & 92.03 $\pm$ 0.06\\
& [0, 32K] &  92.01 $\pm$ 0.05 \\
& [0, 64K] & 92.04 $\pm$ 0.05 \\
& [0, 128K] & 91.64 $\pm$ 0.05 \\
& [0, 256K] & 91.25 $\pm$ 0.03 \\ \cmidrule{2-3}
& [16K, 144K] & 92.08 $\pm$ 0.20 \\
& [32K, 160K] & 92.16 $\pm$ 0.09 \\
& [64K, 192K] & 92.12 $\pm$ 0.04 \\ \midrule
\multirow{3}{*}{ResNet-18 on ImageNet} & None & 67.44 $\pm$ 0.12 \\
& [0, 25] &  66.93 $\pm$ 0.06 \\
& [0, 100] & 66.65 $\pm$ 0.10 \\
\bottomrule
\end{tabular}
\caption{Test accuracy of ResNets trained on CIFAR-10 and ImageNet with a low precision training deficit applied during different windows of time. Deficit windows are listed in terms of training iterations for CIFAR-10 and in terms of epochs for ImageNet.}
\label{tab:img_crit_learn}
\end{small}
\end{table}

\medskip \noindent \textbf{Image Classification.}
We perform similar critical learning period analysis in the image classification domain; see Table \ref{tab:img_crit_learn}. 
On CIFAR-10, we again observe that test accuracy deteriorates smoothly as $R$ increases, reaching a plateau around $R=128K$.
When a deficit window of $128K$ iterations is probed throughout the training process (i.e., bottom CIFAR-10 subsection in Table \ref{tab:img_crit_learn}), we see that the impact of low precision training is most pronounced during the early training iterations.
Due to computational expense, experiments on ImageNet only consider learning deficits applied at the beginning of training.
Nonetheless, we again see that the test accuracy of models trained on ImageNet deteriorates as the value of $R$ is increased.
Even in large-scale experiments, network performance is sensitive to low precision training during the early part of training.

\medskip \noindent \textbf{Discussion.}
Low precision training is a form of learning impairment that can cause permanent performance deterioration if applied during the early phase of learning.
Such behavior is reminiscent of using improper regularization during training, which permanently impairs network performance if applied during early epochs \citep{golatkar2019time}.
Though training precision is known to impact the training process similarly to the learning rate \citep{fu2021cpt}, \textit{it can also be viewed as a form of regularization}, which explains the correlation that is observed between training compute and model performance in Section \ref{S:experiments}.
Although CPT can regularize the training process and improve model performance, schedules that apply quantization too aggressively during the critical period will experience a degradation in performance.
As can be seen in Figure \ref{fig:gnn_clp} and Table \ref{tab:img_crit_learn}, \textit{this problem can be solved by simply delaying the use of low precision until later during the training process.}

\section{Conclusion}
We perform an empirical analysis of different dynamic precision schedules for quantized training with DNNs.
We find for low precision training techniques that a correlation exists between the amount of compute used during training and the model's performance, making the selection of a CPT schedule a simple tool for balancing performance and efficiency in DNN training.
To explain this correlation, we draw a connection between low precision training and critical learning periods, finding that low precision training can permanently deteriorate model performance if applied during early epochs of training.
We hope our findings will be helpful in furthering the adoption of quantized training techniques.










\bibliography{sn-bibliography}

\end{document}